\documentclass{article} 
\usepackage{iclr2019_conference,times}

\usepackage{hyperref}
\usepackage{url}

\usepackage{textcomp}
\usepackage{graphicx}
\usepackage{multirow}
\usepackage{minibox}
\usepackage{xcolor,colortbl}
\usepackage{color}
\usepackage{amsmath}
\usepackage{mathtools}

\usepackage[numbers]{natbib}

\DeclarePairedDelimiter\floor{\lfloor}{\rfloor}

\title{Human Gait Symmetry Assessment using a Depth Camera and Mirrors}

\author{Trong-Nguyen Nguyen\\
DIRO, University of Montreal\\
Montreal, QC, Canada\\
\texttt{nguyetn@iro.umontreal.ca}\\
\And
Huu-Hung Huynh\\
University of Science and Technology\\
Danang, Vietnam\\
\texttt{hhhung@dut.udn.vn}\\
\And
Jean Meunier\\
DIRO, University of Montreal\\
Montreal, QC, Canada\\
\texttt{meunier@iro.umontreal.ca}\\
}

\iclrfinalcopy 
\begin{document}

\maketitle

\begin{abstract}
This paper proposes a reliable approach for human gait symmetry assessment using a depth camera and two mirrors. The input of our system is a sequence of 3D point clouds which are formed from a setup including a Time-of-Flight (ToF) depth camera and two mirrors. A cylindrical histogram is estimated for describing the posture in each point cloud. The sequence of such histograms is then separated into two sequences of sub-histograms representing two half-bodies. A cross-correlation technique is finally applied to provide values describing gait symmetry indices. The evaluation was performed on 9 different gait types to demonstrate the ability of our approach in assessing gait symmetry. A comparison between our system and related methods, that employ different input data types, is also provided.
\end{abstract}

\newcounter{MYtempeqncnt}

\section{Introduction}\label{sec:introduction}

The problem of assessing human gait has received a great attention in the literature since gait analysis is a key component of health diagnosis. Marker-based and multi-camera systems are widely employed to deal with this problem. However, such systems usually require specific equipments with high price tag and sometimes have high computational cost. In order to reduce the cost of devices, we focus on a system of gait analysis which employs only one depth sensor. The principle is similar to a multi-camera system, but the collection of cameras are replaced by one depth sensor and mirrors. Each mirror in our setup plays the role of a camera which captures the scene at a different viewpoint. Since we use only one camera, synchronization can thus be avoided, and the cost of devices is reduced.

In order to simplify the setup, recent studies used a color or depth camera to perform gait analysis. The input of such systems is thus either the subject's silhouette or depth map. Many gait signatures have been proposed based on the former input type such as Gait Energy Image (GEI)~\cite{Han2006}, Motion History Image (MHI)~\cite{Davis2001}, or Active Energy Image (AEI)~\cite{Zhuowen2015}. Typically they are computed based on a side view camera and are usually applied on the problem of human identification. In order to deal with pathological gaits, the input sequence of silhouettes needs more elaborate processing. In the work~\cite{Nguyen2014extracting}, the input sequence of silhouettes was separated into consecutive sub-sequences corresponding to gait cycles. The feature extraction was applied on each individual silhouette and the gait assessment was performed based on a combination of such features in each sub-sequence. Instead of capturing a side view of the subject, the authors in~\cite{Bauckhage2005,Bauckhage2009} put the camera in front of a walking person and tried to detect unusual movement. The movement of the subject was encoded based on a sequence of lattices applied on the captured silhouettes. A feature vector was then estimated for each lattice according to a predefined set of points, and the characteristic representing the whole motion was formed by concatenating such vectors. This step of concatenation is to incorporate the temporal context into the classification with a Support Vector Machine (SVM). A common limitation of such silhouette-based approaches is the reduction of data dimension since the 3D scene is represented by 2D images. In order to overcome this drawback, a depth camera is often employed. One of the devices that are widely used is the Microsoft Kinect. Beside its low price, this camera provides a built-in functionality of human skeleton localization, estimated in each single depth frame~\cite{Shotton2011realtime,Shotton2013efficient}. Such skeletal information is useful for gait-related problems such as abnormal gait detection~\cite{Nguyen2016}, gait-based recognition~\cite{Jiang2015}, and pathological gait analysis~\cite{Bigy2015}. A limitation of skeleton-based approaches is that the skeleton may be deformed due to self-occlusions in the depth map. Unfortunately, such problem usually occurs in pathological gaits~\cite{Pfister2014,Auvinet2015}. 
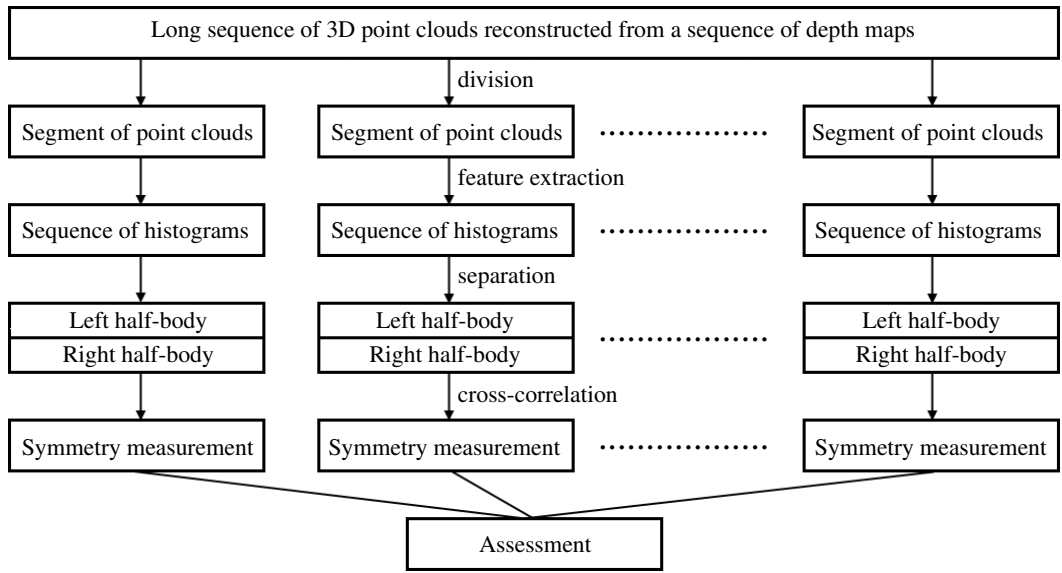
\begin{figure*}[t]
\centering
\footnotesize
\scalebox{1}{
\begin{picture}(404,210)
	\put(0,0){\includegraphics[width=\textwidth]{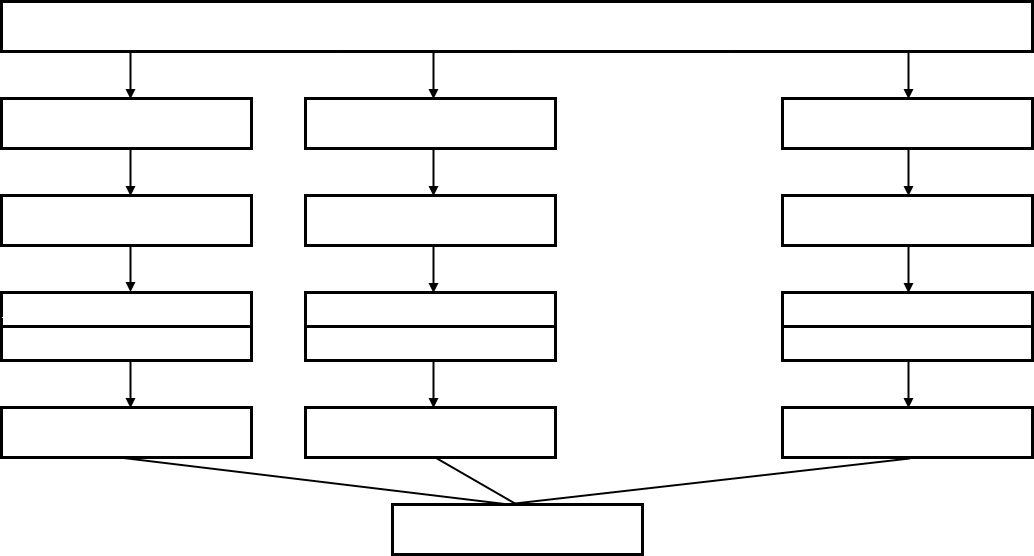}}
		\put(54,202){Long sequence of 3D point clouds reconstructed from a sequence of depth maps}
		\put(170,183){division}
		\put(5,163){Segment of point clouds}
		\put(122,163){Segment of point clouds}
		\put(304,163){Segment of point clouds}
		\put(223,165){\Large..................}
		\put(170,146){feature extraction}
		\put(5,126){Sequence of histograms}
		\put(122,126){Sequence of histograms}
		\put(305,126){Sequence of histograms}
		\put(223,128){\Large..................}
		\put(170,109){separation}
		\put(23,92){Left half-body}
		\put(20.5,79){Right half-body}
		\put(139,92){Left half-body}
		\put(136.5,79){Right half-body}
		\put(323,92){Left half-body}
		\put(320.5,79){Right half-body}
		\put(223,87){\Large..................}
		\put(170,64){cross-correlation}
		\put(5,44){Symmetry measurement}
		\put(121,44){Symmetry measurement}
		\put(305,44){Symmetry measurement}
		\put(223,46){\Large..................}
		\put(178,7){Assessment}
\end{picture}}
\caption{Flowchart of our processing.}
\label{fig:overview}
\end{figure*}
For that reason, other researchers have used depth images without skeleton fitting to assess gait. For instance,~\citeauthor{Auvinet2015}~\cite{Auvinet2015} have proposed an asymmetry index based on spatial differences between the lower limb positions recorded by a depth camera placed in front of the subject walking on a treadmill. They have demonstrated that this approach was more reliable than a similar index based on skeleton information. However these differences were computed only in a small predefined leg zone, required averaging depth maps over several gait cycles (to be determined) and limited to a single depth map representing only a partial view of the whole body.~\citeauthor{Nguyen2018BHI}~\cite{Nguyen2018BHI} have also employed successfully (enhanced) depth maps for gait assessment using a weighted combination of a PoI-score, based on depth map key points, and a LoPS-score describing a measurement of body balance from the body silhouette. However, their method was still limited to a partial view of the body and basic features.

Taking all this into account, we present an original approach that estimates an index of human gait symmetry without requiring skeleton extraction or gait cycle detection. To improve the performance, the input of our system is a sequence of 3D point clouds of the whole body obtained with a combination of a depth camera and two mirrors.  Cylindrical histograms of point cloud are then computed and analysed for left-right symmetry for subjects walking on a treadmill to obtain their symmetry index. The remaining of this paper is organized as follow: Section~\ref{sec:method} describes details of our method including the setup, point cloud formation, feature extraction, and gait symmetry assessment; our experiments and evaluation are presented in Section~\ref{sec:experiment}, and Section~\ref{sec:conclusion} gives the conclusion.

\section{Proposed method}\label{sec:method}
In order to give a visual understanding, an overview of the proposed approach is shown in Fig.~\ref{fig:overview}. 

\subsection{Point cloud formation}

Beside a ToF depth camera and two mirrors, our setup also employs a treadmill where each subject performs his/her walking gait. The ToF camera is put in front of the subject and the two mirrors are behind so that the walking person nearly stands at the center (see Fig.~\ref{fig:setupsketch}). An example of such captured depth map is presented in Fig.~\ref{fig:depthmap}.
\begin{figure}[t]
\centering
\begin{picture}(250,240)
	\put(20,0){\includegraphics[scale=0.8]{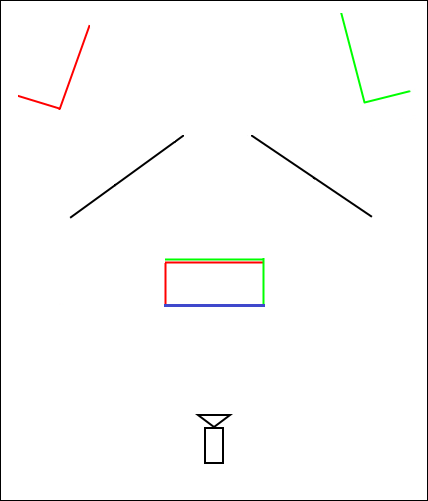}}
	\put(59,149){\rotatebox{37}{mirror 1}}
	\put(160,170){\rotatebox{-37}{mirror 2}}
	\put(94,5){depth camera}\put(110,102){object}
\end{picture}
\caption{Basic principle of the depth camera system with mirrors. The depth information visible by the depth camera (blue surface of the object) is complemented by the reflected depth information from the two mirrors (red and green surfaces) to obtain the full 3D reconstruction of the object. Notice that in practice some unreliable points must be removed due to multiple reflections with ToF camera (see~\cite{NguyenKinect2,NguyenReportKinect2}).}
\label{fig:setupsketch}
\end{figure}
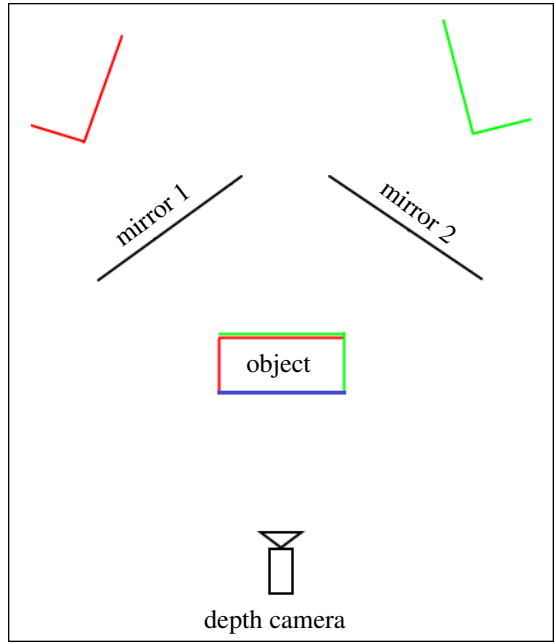
\begin{figure}[t]
\centering
	\includegraphics[width=0.64\textwidth]{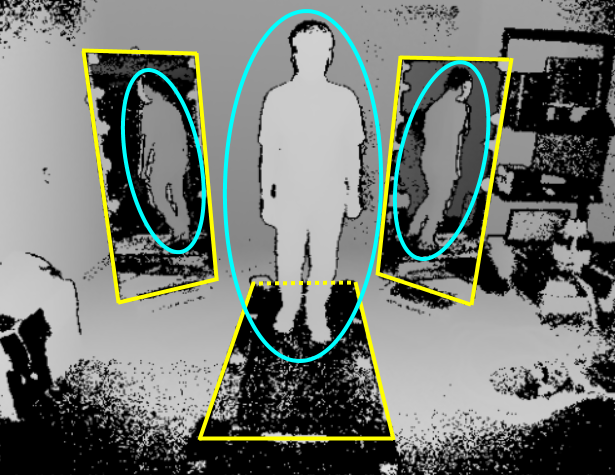}
\caption{A depth map captured by our system, in which there are 3 collections of subject's pixels (highlighted by cyan ellipses). The two mirrors and the treadmill are highlighted with yellow rectangles.}
\label{fig:depthmap}
\end{figure}

There are two popular types of depth sensor that are distinguished based on the scheme of depth estimation: structured light (SL) and Time-of-Flight (ToF). In our work, the second type was employed because it is more accurate~\cite{Wasenmuller2017} and consequently its point cloud has a higher level of details compared with the first one.

As shown in Fig.~\ref{fig:depthmap}, each captured depth map provides subject's images from 3 different view points. The 3D reconstruction could also be performed when the depth camera is replaced by a color one. However, the process of reconstruction based on such data produces an object (visual hull) that is bigger, less accurate and contains redundancies as described in~\cite{Auvinet2012}. Therefore employing a depth camera in our setup is advantageous to provide a better model of 3D information.

\begin{figure}[t]
\centering
	\includegraphics[width=0.7\textwidth]{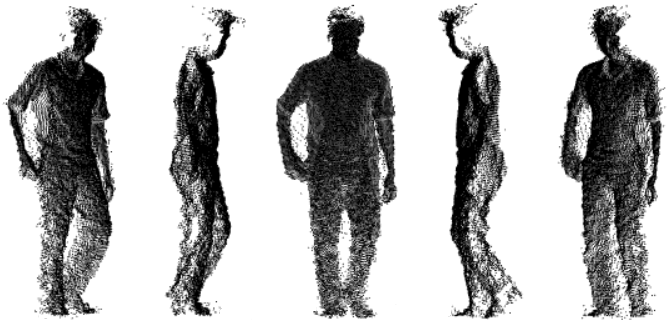}
\caption{A point cloud obtained in our setup seen from different view points.}
\label{fig:cloudexample}
\end{figure}

Let us briefly describe the formation of a 3D point cloud from each depth map captured by a depth camera in our work. According to the example shown in Fig.~\ref{fig:depthmap}, a depth map contains 3 partial surfaces of the subject. A point cloud representing the walking person can thus be formed by combining (a) the direct cloud (highlighted by the middle ellipse) and (b) reflections of two indirect ones (smaller ellipses), which are behind the mirrors~\cite{Nguyen2018SPIE,NguyenKinect2}. We used the method described in~\cite{NguyenKinect2,NguyenReportKinect2} because it was specifically designed for ToF camera and is robust to unreliable points caused by unwanted multiple reflections. Figure~\ref{fig:cloudexample} illustrates an example of a 3D point cloud obtained with the setup of Fig.~\ref{fig:depthmap}.

\subsection{Feature extraction}

In order to perform gait symmetry assessment, we separate the entire movement into two non-overlapping sub-movements corresponding to the left and right half-bodies. In more details, each individual point cloud is processed to obtain a cylindrical histogram, and then the histogram is split into two sub-histograms representing two half-bodies.

\subsubsection{Coordinate system transformation}
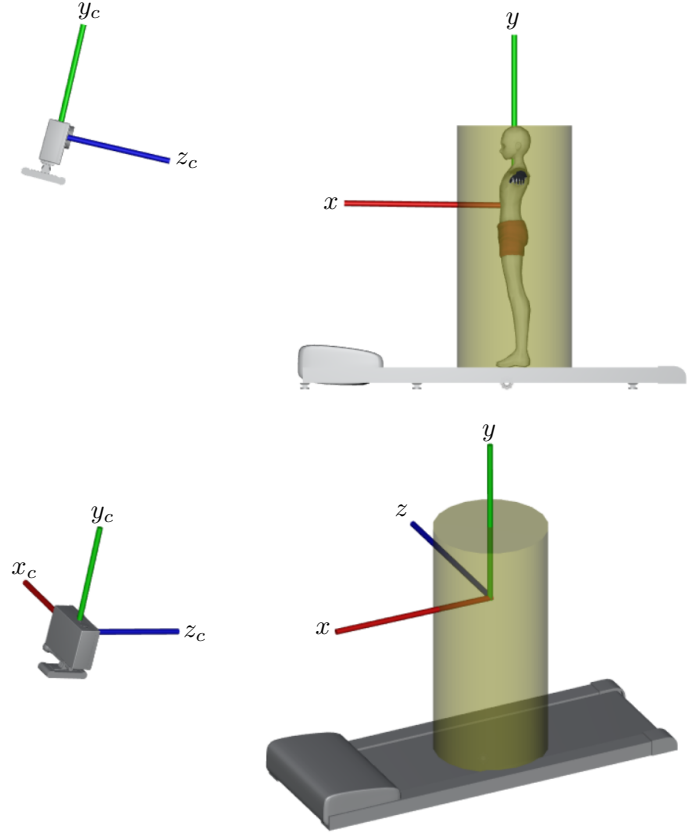
\begin{figure}[t]
\centering
\begin{picture}(250,314)
	\put(0,167){\includegraphics[scale=0.465]{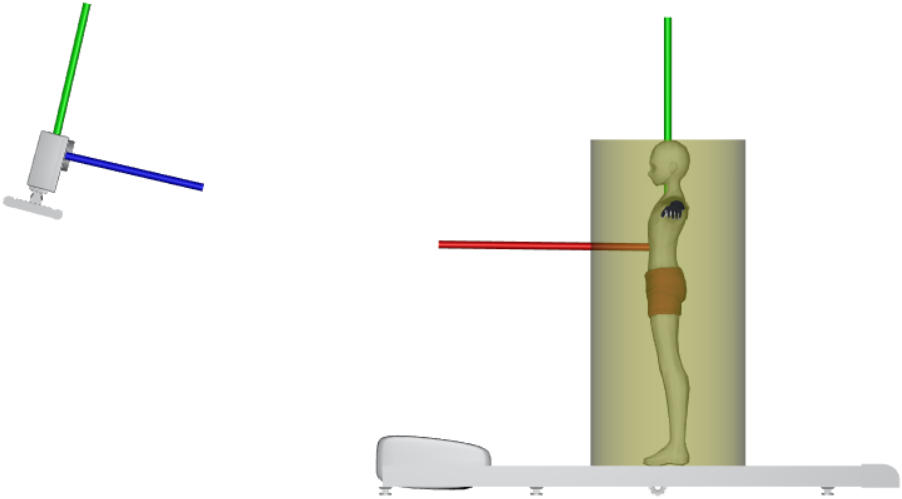}}
	\put(0,0){\includegraphics[scale=0.49]{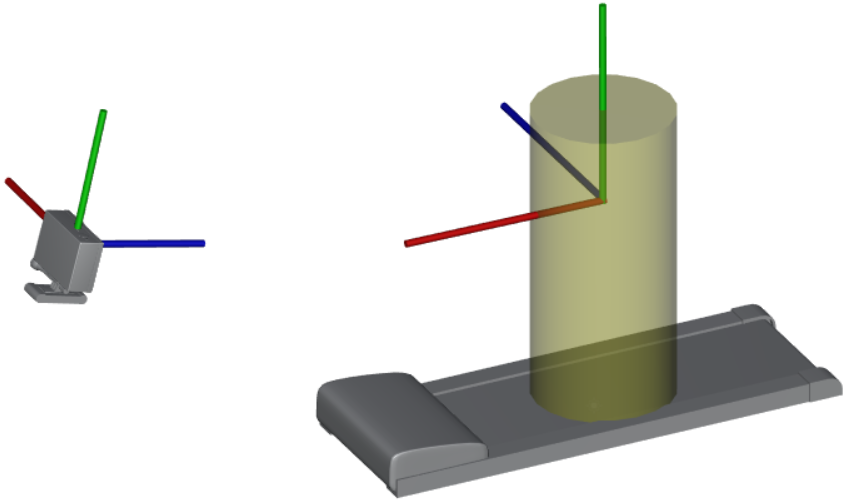}}
	\put(22,310){$y_c$}\put(59,253){$z_c$}
	\put(184,306){$y$}\put(115,236){$x$}
	\put(-3,98){$x_c$}\put(27,119){$y_c$}\put(62,74){$z_c$}
	\put(112,75){$x$}\put(175,150){$y$}\put(142,120){$z$}
\end{picture}
\caption{Visualizations of our scene from two different view points that show the camera coordinate system and the body coordinate system used for matching a cylinder with a point cloud. They are right-handed.}
\label{fig:rigidtransform}
\end{figure}

Let us notice that the point cloud is initially computed in the camera space $(x_c, y_c, z_c)$. Therefore, to facilitate the computation of the cylindrical histogram, we need a rigid transformation from the camera coordinate system to the object (body) coordinate system. The latter is defined by its origin assigned to the centroid of the body 3D point cloud, the $y$-axis normal to the ground (treadmill), the $x$-axis along the walking direction and the $z$-axis in the left to right direction (see Fig.~\ref{fig:rigidtransform}). The $y$-axis is easily estimated as the normal to the treadmill plane obtained during calibration using a few markers (4 in our work). The walking direction ($x$-axis) is determined from the vector between two appropriate markers on the treadmill. The remaining dimension ($z$-axis) is estimated by performing a cross product.

\subsubsection{Cylindrical histogram estimation}
Once the subject's point cloud corresponding to each depth frame has been transformed, its symmetrical characteristic is then extracted with a cylindrical histogram. Concretely, a cylinder is estimated with the main axis coinciding with the $y$-axis of the body coordinate system, and the top and bottom surfaces going through the highest and lowest points along this dimension. The cylinder's radius is long enough to guarantee that the entire point cloud is within the cylinder.

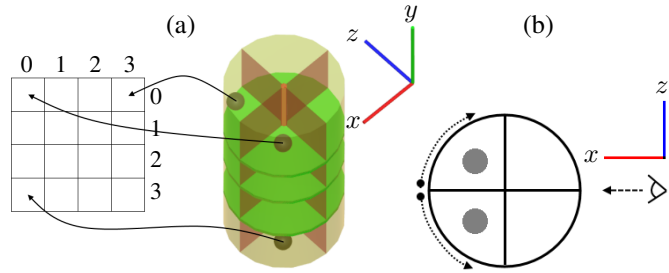
\begin{figure}[t]
\centering
\begin{picture}(250,105)
	\put(0,0){\includegraphics[scale=0.36]{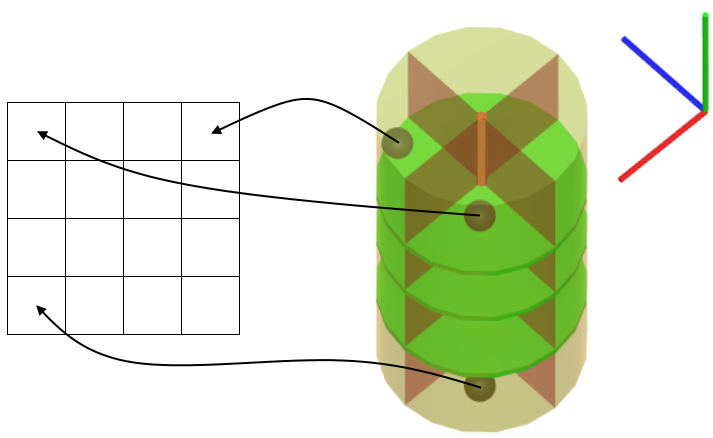}}
	\put(154,0){\includegraphics[scale=0.55]{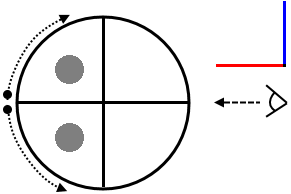}}
	\put(128,54){$x$}\put(150,96){$y$}\put(128,87){$z$}
	\put(218,41){$x$}\put(245,67){$z$}
	\put(60,90){(a)}\put(195,90){(b)}
	\put(5,75){\small{0}}\put(18,75){\small{1}}\put(31,75){\small{2}}\put(44,75){\small{3}}
	\put(54,63){\small{0}}\put(54,51){\small{1}}\put(54,39){\small{2}}\put(54,27){\small{3}}
\end{picture}
\caption{Mapping from cylindrical sectors to histogram's bins. The sub-figure (a) shows a 3D visualization. The histogram can be considered as a flattened cylinder seen from a specific view point as the sub-figure (b). In this simplified representation, the histogram's size is $4 \times 4$ corresponding to 16 sectors.}
\label{fig:correspondance}
\end{figure}

Given a cloud $P$ of $n$ 3D points and the size $h \times w$ of a target cylindrical histogram (see Fig.~\ref{fig:correspondance}), the sector's zero-based index of each point $P^{(i)}$ is determined as
\begin{equation}
	\begin{cases}h^{(i)}=min\big(\floor*{h(max_y-P^{(i)}_y)(max_y-min_y)^{-1}},h-1\big)\\w^{(i)}=\floor*{\frac{w}{2\pi}\{[2\pi+sgn(\vec{v}^{(i)}_z)cos^{-1}(\frac{\vec{v}^{(i)}_x}{\|\vec{v}^{(i)}\|})]\bmod (2\pi)\}}\end{cases} 
	\label{eq:sectorindex}
\end{equation}
where $max_y$ and $min_y$ respectively indicates the $y$-coordinate of highest and lowest points in the cloud $P$ along the $y$-axis, $\lfloor \circ \rfloor$ is the floor function, $P^{(i)}_y$ is the $y$ value of point $P^{(i)}$, $sgn(\circ)$ is the sign function, and $\vec{v}^{(i)}$ is a 2D vector computed from the $y$-axis to the point $P^{(i)}$.
Notice that the notation $\vec{v}^{(i)}_z$ in eq.~(\ref{eq:sectorindex}) is the $z$ coordinate of $\vec{v}^{(i)}$. The subscript $z$ is to indicate the axis used in this calculation. The $min$ function in (\ref{eq:sectorindex}) is to guarantee that the output index is in the range $[0, h-1]$.

\begin{figure}[t]
\centering
\begin{picture}(250,128)
	\put(0,0){\includegraphics[scale=0.55]{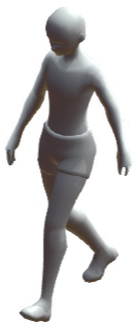}}
	\put(45,0){\includegraphics[scale=0.55]{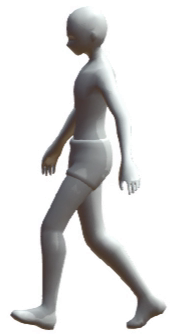}}
	\put(100,69.7){\includegraphics[scale=0.1]{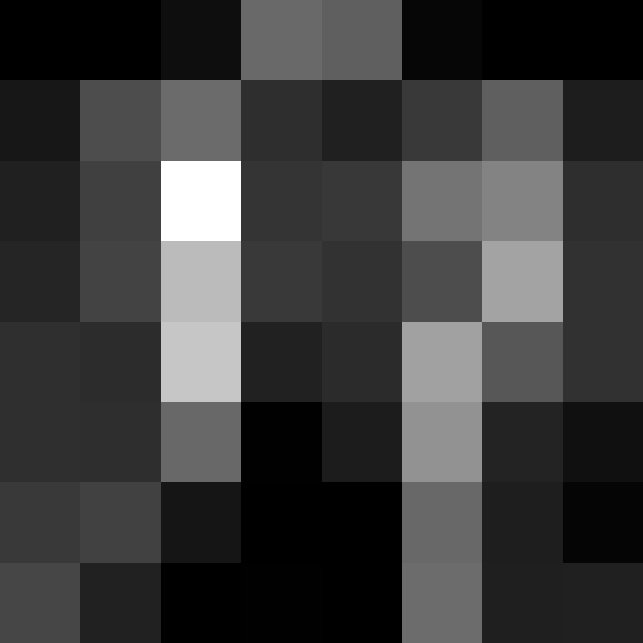}}
	\put(142,0){\includegraphics[scale=0.28]{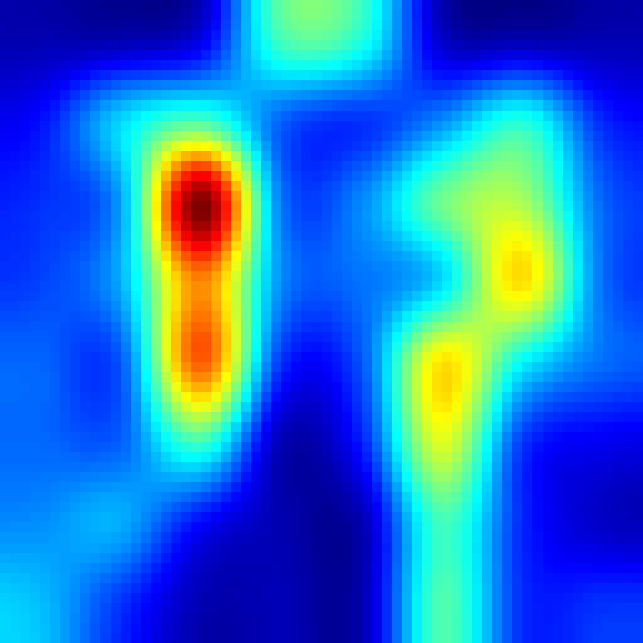}}
	\put(186,120){head}\put(196,118){\line(0,-1){15}}
\end{picture}
\caption{Example of flattened cylindrical histogram. The original histogram (gray image) of size $8 \times 8$ is scaled and is represented as a heat map for a better visualization. We can explicitly see the posture's self-symmetry since the head is at the center of the histogram.}
\label{fig:histseparation}
\end{figure}
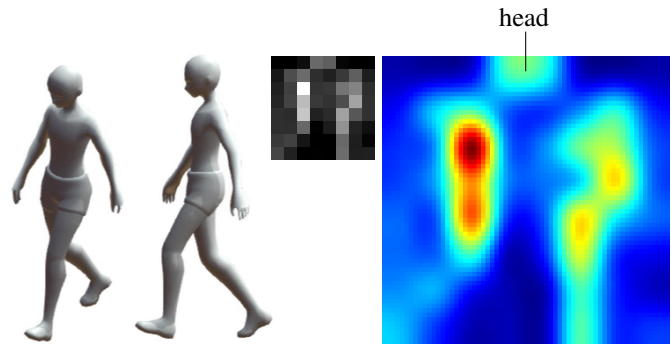

Although a cylinder is employed to estimate a histogram for each point cloud, the representation of such histogram is flat, i.e. a matrix of size $h \times w$. The correspondence between a histogram's bin and its original cylinder's sector is illustrated in Fig.~\ref{fig:correspondance}. As illustrated in Fig.~\ref{fig:histseparation}, the head is aligned at the center of the cylindrical histogram after performing the estimation. Notice that a slight \textit{rotation} of the cylinder might be necessary to ensure that the body is well centered in the cylindrical histogram depending on the camera-to-body rigid transformation accuracy (see above).

\subsection{Gait symmetry assessment}\label{sec:assessment}

\begin{figure*}[t]
\centering
\footnotesize
\begin{picture}(404,150)
	\put(0,10){\includegraphics[width=\textwidth]{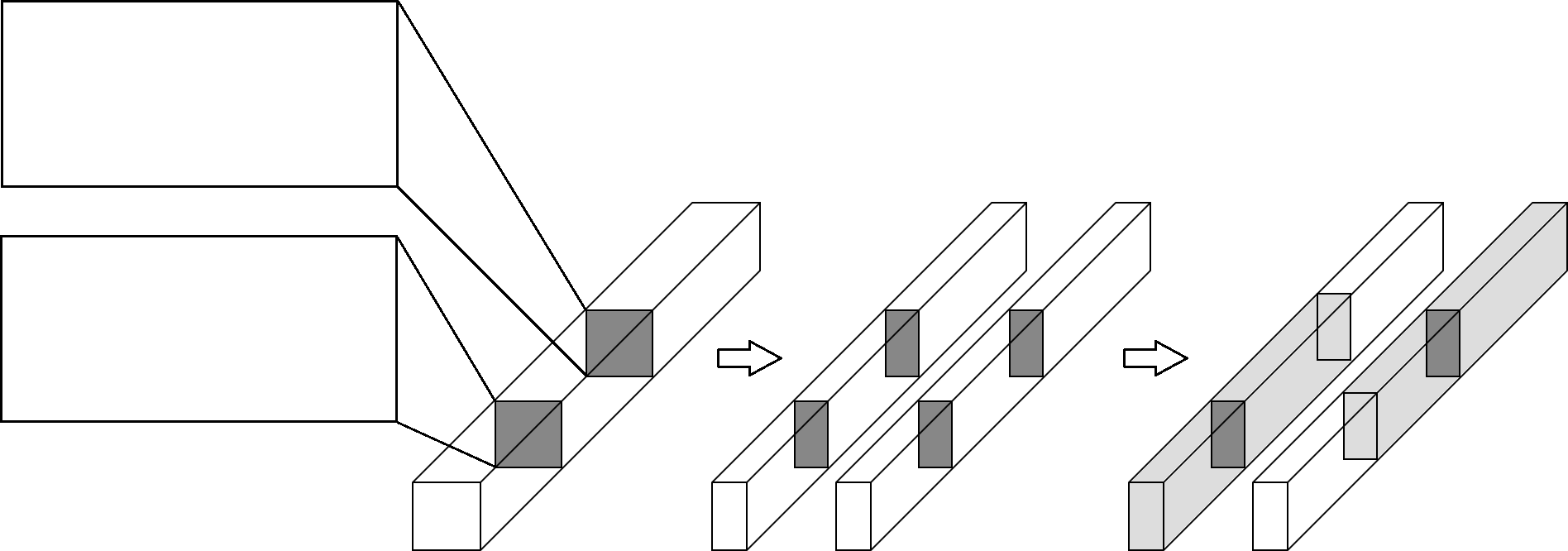}}
	\put(291,65){\rotatebox{47}{cross-correlation}}
	\put(195,87){$h$}\put(180,102){$w$}
	\put(262,87){$h$}\put(247,102){$w/2$}\put(294,87){$h$}\put(279,102){$w/2$}
	\put(143.5,40){$i^{th}$}\put(167,64){$j^{th}$}
	\put(53,0){(a) Sequence of histograms}
	\put(158,0){(b) Half-body sequences}
	\put(255,0){(c) Best matching of different shiftings}
	\put(58,46){\includegraphics[scale=0.105]{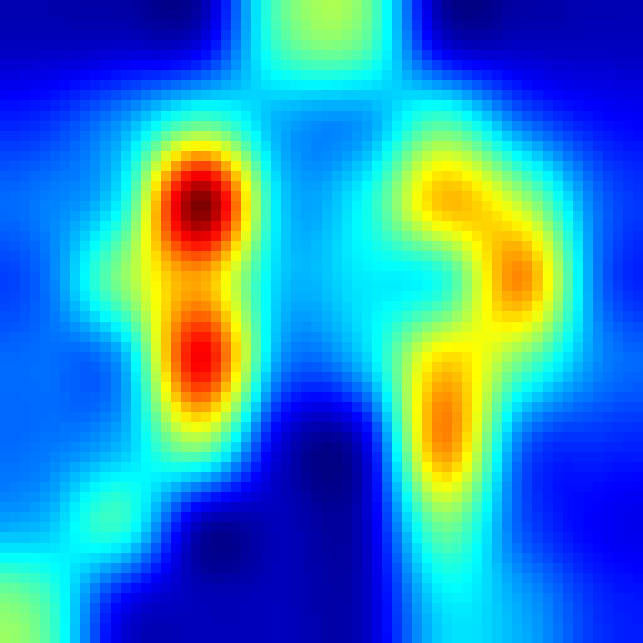}}
	\put(3,46){\includegraphics[scale=0.21]{images/gait32a.png}}
	\put(27,46){\includegraphics[scale=0.21]{images/gait32b.png}}
	\put(58,105){\includegraphics[scale=0.105]{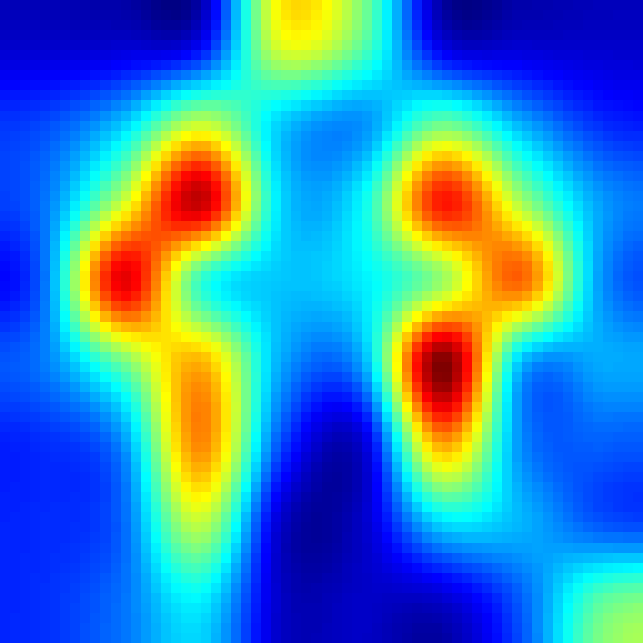}}
	\put(3,105){\includegraphics[scale=0.21]{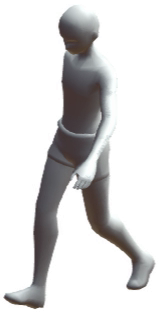}}
	\put(27,105){\includegraphics[scale=0.21]{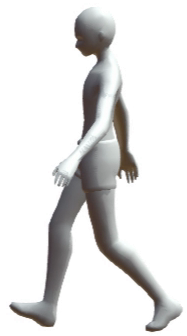}}
\end{picture}
\caption{Symmetry assessment for a sequence of histograms. We say that the $i^{th}$ and $j^{th}$ histograms have a good symmetry since each one and the horizontal flip version of the other are quite similar. The heat maps in this figure are enhanced (for visualization) from actual histograms estimated from 3D point clouds in our experiments, and the 3D models are used for illustrating the corresponding postures. Instead of performing the cross-correlation on the input sequence and its clone, we process directly on two sequences corresponding to half-bodies to reduce the number of calculations and memory requirement. Notice that an input sequence may contain similar histograms because walking is a periodic motion.}
\label{fig:shifting}
\end{figure*}

Similarly to related studies on gait analysis (e.g.~\cite{Bauckhage2009,Nguyen2016,Auvinet2015,Nguyen2018BHI}), the assessment of gait symmetry in our system also considers the temporal factor. Concretely, the value measuring the gait symmetry is estimated on a sequence of consecutive histograms. Symmetry can be measured by separating a histogram into two sub-histograms corresponding to two half-bodies. In other words, a sequence of histograms of size $h \times w$ becomes two sequences of sub-histograms of size $h \times 0.5w$. According to the nature of normal walking gait, there is a shifting along the time axis between a left sub-histogram and its corresponding symmetric right one. Therefore our method employs a cross-correlation technique~\cite{Stoica2005} to measure the gait symmetry index. A good symmetry occurs if the left sub-histogram is similar to the horizontal flip version of the (shifted) right sub-histogram (Fig.~\ref{fig:shifting}).
\begin{figure*}[!t]
\normalsize
\setcounter{MYtempeqncnt}{\value{equation}}
\setcounter{equation}{1}
\begin{equation}
	S \big(L,R,D\big) =min \big(\big\{\frac{1}{l-|d|} \sum_{i=0}^{l-|d|-1} Diff \big(L_{max(0,d)+i},R^f_{max(0,-d)+i}\big)   \mid d \in D \big\} \big)
	\label{eq:crosscorrelation}
\end{equation}
\setcounter{MYtempeqncnt}{\value{equation}}
\setcounter{equation}{\value{MYtempeqncnt}}
\hrulefill
\vspace*{4pt}
\end{figure*}
\begin{figure}[t]
\centering
\footnotesize
\begin{picture}(250,126)
	\put(0,0){\includegraphics[width=0.64\textwidth]{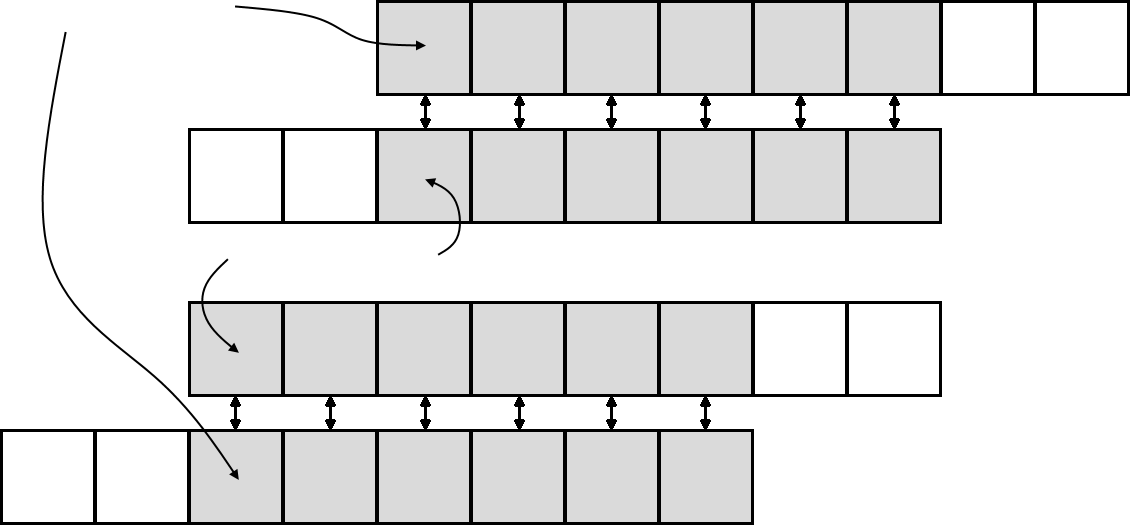}}
	\put(215,38){$Ref$}\put(215,75){$Ref$}
	\put(3,25){$d=-2$}\put(51,94){$d=2$}
	\put(55,58){$max(0,d)$}
	\put(5,115){$max(0,-d)$}
\end{picture}
\caption{Correlation between two sequences corresponding to positive and negative shifting values $d$, and indices of beginning positions. The notation $Ref$ indicates the reference (left sequence in our work). In these two examples, the lengths of each input sequence and the common one are 8 and 6, respectively.}
\label{fig:shiftdirection}
\end{figure}

The processing of this stage is as follows. The input is a sequence of histograms. Although many related studies tried to process on gait cycles, our assessment is performed on consecutive (i.e. non-overlapping) sub-sequences (or segments) that have the same length. There are several reasons leading to our choice: (1) gait cycle determination would be difficult to perform when working on pathological gaits, (2) the symmetry can be measured well by dealing with the mentioned shifting on an arbitrary (long enough) sequence of histograms, and (3) sub-sequences do not need to have common properties (e.g. similar beginning and ending postures as in~\cite{Nguyen2016} or~\cite{Auvinet2015}) because we do not focus on training a model representing the gait. Each sub-sequence is then separated into two sequences of left and right sub-histograms. We can expect that by assigning a sequence as the reference and shifting the other with an appropriate delay, the two registered sub-sequences would have a good symmetry (see Fig.~\ref{fig:shifting}). Because such suitable delay is various with different subjects, we perform the shifting with a set of delays and choose the best match. Given two sequences of sub-histograms $L$ and $R^f$ ($R$ horizontally flipped) of length $l$ representing two half-bodies, a set of shifting delays $D$, the symmetry index $S$ is measured as in eq.~(\ref{eq:crosscorrelation}). The $Diff$ function estimates the distance between two sub-histograms, the $min$ function thus provides the best matching. Notice that the left segment is assigned as the reference, and the set $D$ contains both negative and positive values indicating the shifting direction of the other segment (see Fig.~\ref{fig:shiftdirection}). At the end of this stage, the system provides a sequence of scores measuring the gait symmetry corresponding to consecutive segments.

\section{Experimental results}\label{sec:experiment}

\subsection{Data acquisition}\label{sec:acquisition}
Our experiments were performed on 9 different gait types consisting of normal walking gaits and 8 simulated asymmetrical (so-called abnormal) ones. These abnormal gaits were simulated by either padding a sole with a thickness of 5/10/15-centimeters under one foot or attaching a weight (4 kilograms) to one ankle. We use the notations L$|$5cm, L$|$10cm, L$|$15cm, and L$|$4kg to indicate these abnormal gaits with left leg, and so on for the other leg. Such set up can provide gaits having a higher level of asymmetry compared with normal walking ones. A Kinect 2 was employed for data acquisition since it uses ToF for depth measurement and had a low price. There were 9 volunteers that performed the 9 mentioned walking gaits, in which each motion was captured as 1200 continuous frames with a frame rate of 13 fps. The treadmill speed was set at 1.28 km/h. In order to provide a comparison with related approaches, we also captured other data types including skeleton and silhouette using built-in functionalities of the Kinect 2. Therefore each walking gait of a volunteer is represented by 1200 point clouds, 1200 skeletons, and 1200 silhouettes~\cite{NguyenReportDataset}\footnote{\url{http://www.iro.umontreal.ca/~labimage/GaitDataset}}\textsuperscript{,}\footnote{Prepared cylindrical histograms of size $16 \times 16$ are available at \url{http://www.iro.umontreal.ca/~labimage/GaitDataset/rawhists.zip} (unzip password: \textbf{8sxDaUEmUG})}. These experimental procedures involving human subjects were approved by the Institutional Review Board (IRB).

\subsection{System parameters}
As mentioned in Section~\ref{sec:assessment}, the input sequence of point clouds is segmented into non-overlapping segments. In our experiments, each input sequence was separated into 10 segments of length 120 (about 9 seconds), the corresponding output was thus a vector of 10 elements measuring the gait symmetry. The size of cylindrical histogram was $16 \times 16$, so each half-body volume in Fig.~\ref{fig:shifting} had a size of $[16 \times 8 \times 120]$. The $L_1$ norm was used for measuring the distance [the term $Diff$ in eq.~(\ref{eq:crosscorrelation})] between two normalized histograms, i.e. dividing each bin value by the sum. The shifting delays were in the range $[-50, 50]$ to guarantee that the length of the common sub-sequence would be greater than a half of input length. Let us notice that $16 \times 16$ is \textit{not} the optimal size of cylindrical histograms. This is just an arbitrarily selected value for our experiments. The effect of that hyperparameter will be discussed in Section~\ref{sec:sizeaffection}.

\subsection{Testing results}\label{sec:ourresult}
\begin{figure}[t]
\centering
\begin{picture}(250,235)
	\put(19,3){\includegraphics[scale=0.53]{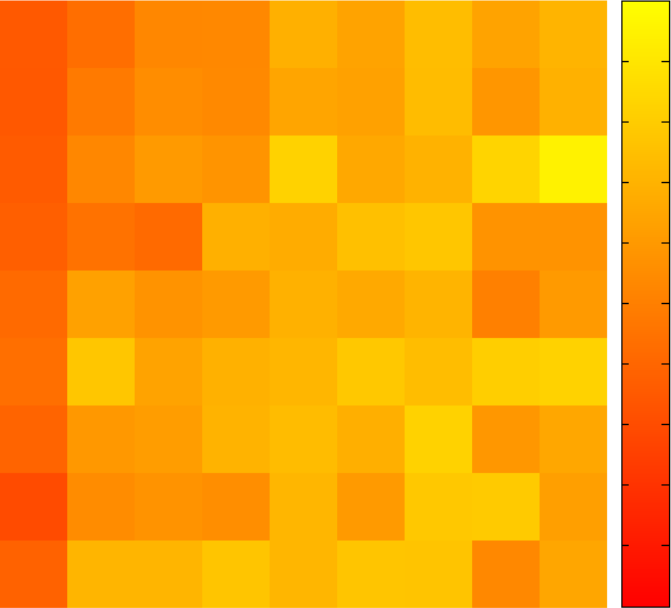}}
	\put(238,0){\small 0} \put(238,20){\small 0.1}
	\put(238,39.5){\small 0.2} \put(238,59){\small 0.3}
	\put(238,78.5){\small 0.4} \put(238,98){\small 0.5}
	\put(238,118){\small 0.6} \put(238,136.5){\small 0.7}
	\put(238,156){\small 0.8} \put(238,175){\small 0.9}
	\put(238,195){\small 1}
	\put(5,11){$v_9$}\put(5,33){$v_8$}\put(5,55){$v_7$}
	\put(5,78){$v_6$}\put(5,100){$v_5$}\put(5,122){$v_4$}
	\put(5,143){$v_3$}\put(5,165){$v_2$}\put(5,187){$v_1$}
	\put(40,206){\includegraphics[scale=0.38]{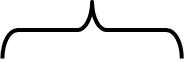}}
	\put(83,206){\includegraphics[scale=0.38]{images/bracket.png}}
	\put(126,206){\includegraphics[scale=0.38]{images/bracket.png}}
	\put(169,206){\includegraphics[scale=0.38]{images/bracket.png}}
	\put(52,225){5cm}
	\put(93,225){10cm}
	\put(136,225){15cm}
	\put(183,225){4kg}
	\put(25,200){N}
	\put(47,200){L}\put(69,200){R}\put(91,200){L}\put(113,200){R}
	\put(135,200){L}\put(156,200){R}\put(177,200){L}\put(198,200){R}
	\put(22,185){\small .35}\put(22,163){\small .35}\put(22,141){\small .36}
	\put(22,120){\small .37}\put(22,98){\small .41}\put(22,76){\small .44}
	\put(22,53){\small .39}\put(22,31){\small .30}\put(22,9){\small .39}
	\put(44,185){\small .43}\put(44,163){\small .48}\put(44,141){\small .53}
	\put(44,120){\small .45}\put(44,98){\small .63}\put(44,76){\small .77}
	\put(44,53){\small .60}\put(44,31){\small .55}\put(44,9){\small .71}
	\put(66,185){\small .53}\put(66,163){\small .55}\put(66,141){\small .60}
	\put(66,120){\small .42}\put(66,98){\small .58}\put(66,76){\small .64}
	\put(66,53){\small .61}\put(66,31){\small .58}\put(66,9){\small .71}
	\put(88,185){\small .53}\put(88,163){\small .54}\put(88,141){\small .58}
	\put(88,120){\small .69}\put(88,98){\small .61}\put(88,76){\small .69}
	\put(88,53){\small .70}\put(88,31){\small .56}\put(88,9){\small .77}
	\put(110,185){\small .69}\put(110,163){\small .65}\put(110,141){\small .82}
	\put(110,120){\small .67}\put(110,98){\small .69}\put(110,76){\small .72}
	\put(110,53){\small .74}\put(110,31){\small .72}\put(110,9){\small .71}
	\put(132,185){\small .64}\put(132,163){\small .63}\put(132,141){\small .66}
	\put(132,120){\small .75}\put(132,98){\small .66}\put(132,76){\small .78}
	\put(132,53){\small .69}\put(132,31){\small .60}\put(132,9){\small .77}
	\put(153,185){\small .74}\put(153,163){\small .74}\put(153,141){\small .70}
	\put(153,120){\small .77}\put(153,98){\small .70}\put(153,76){\small .74}
	\put(153,53){\small .82}\put(153,31){\small .78}\put(153,9){\small .77}
	\put(174,185){\small .64}\put(174,163){\small .59}\put(174,141){\small .83}
	\put(174,120){\small .58}\put(174,98){\small .50}\put(174,76){\small .81}
	\put(174,53){\small .59}\put(174,31){\small .79}\put(174,9){\small .53}
	\put(195,185){\small .71}\put(195,163){\small .69}\put(195,141){\small .95}
	\put(195,120){\small .58}\put(195,98){\small .61}\put(195,76){\small .82}
	\put(195,53){\small .65}\put(195,31){\small .62}\put(195,9){\small .65}
\end{picture}
\caption{Mean values of 10 measurements provided by our system for each gait of each volunteer. The notation N indicates normal gaits, L and R respectively represent left and right legs, and $v_i$ is the $i^{th}$ volunteer.}
\label{fig:overallresult}
\end{figure}

Since our system returned 10 measurement values (corresponding to 10 segments of length 120) for each input sequence of point clouds, their mean can be used as an index of gait symmetry. The experimental results are shown in Fig.~\ref{fig:overallresult}.
\begin{figure*}[t]
\centering
\scalebox{0.76}{
\begin{picture}(520,375)
	\put(125,334){\framebox(267,34)}
	\put(130,337){\includegraphics[scale=0.45]{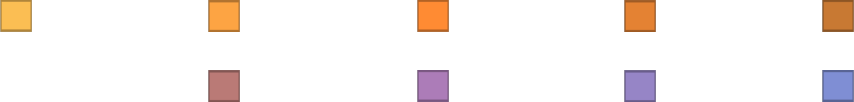}}
	\put(142,358){\small normal}\put(198,358){\small L$|$5cm}
	\put(254,358){\small L$|$10cm}\put(310,358){\small L$|$15cm}
	\put(364,358){\small L$|$4kg}\put(198,339){\small R$|$5cm}
	\put(254,339){\small R$|$10cm}\put(310,339){\small R$|$15cm}
	\put(364,339){\small R$|$4kg}
	\put(0,220){\includegraphics[scale=0.28]{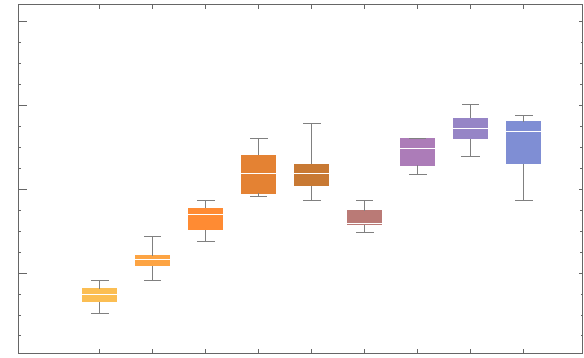}}
	\put(105,230){volunteer 1}
	\put(-3,241){\footnotesize .4}\put(-3,265){\footnotesize .6}
	\put(-3,289){\footnotesize .8}\put(-1,313){\small 1}
	\put(175,220){\includegraphics[scale=0.28]{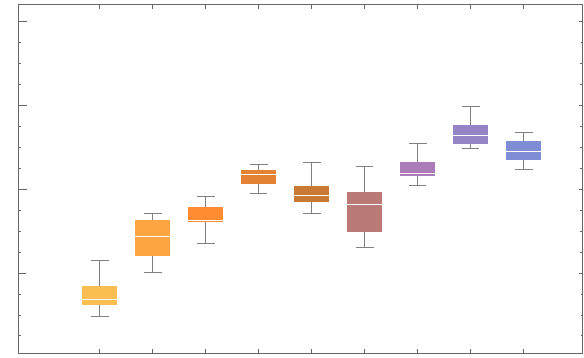}}
	\put(280,230){volunteer 2}
	\put(172,241){\footnotesize .4}\put(172,265){\footnotesize .6}
	\put(172,289){\footnotesize .8}\put(174,313){\small 1}
	\put(350,220){\includegraphics[scale=0.28]{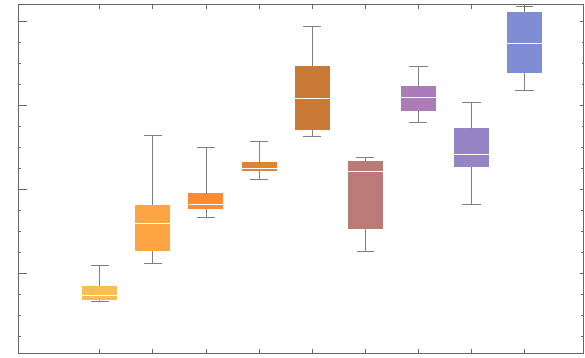}}
	\put(455,230){volunteer 3}
	\put(347,241){\footnotesize .4}\put(347,265){\footnotesize .6}
	\put(347,289){\footnotesize .8}\put(349,313){\small 1}
	\put(0,110){\includegraphics[scale=0.28]{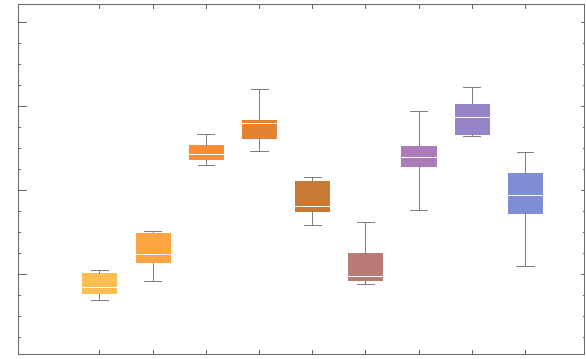}}
	\put(105,120){volunteer 4}
	\put(-3,131){\footnotesize .4}\put(-3,155){\footnotesize .6}
	\put(-3,179){\footnotesize .8}\put(-1,203){\small 1}
	\put(175,110){\includegraphics[scale=0.28]{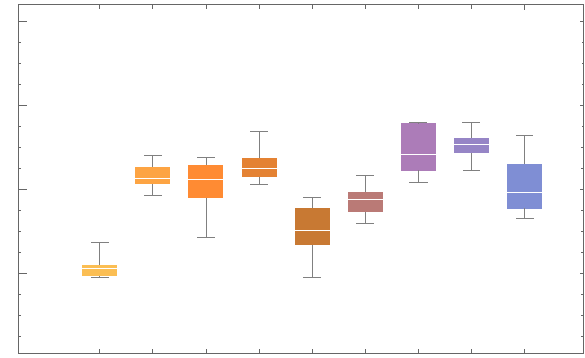}}
	\put(280,120){volunteer 5}
	\put(172,131){\footnotesize .4}\put(172,155){\footnotesize .6}
	\put(172,179){\footnotesize .8}\put(174,203){\small 1}
	\put(350,110){\includegraphics[scale=0.28]{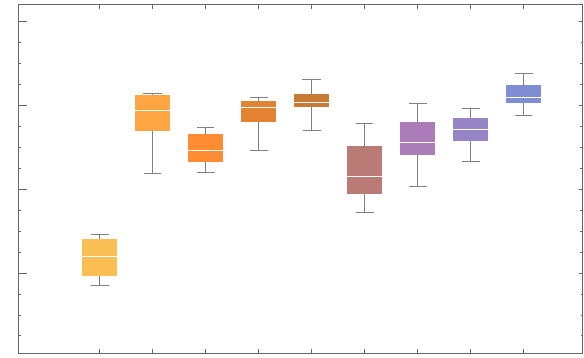}}
	\put(455,120){volunteer 6}
	\put(347,131){\footnotesize .4}\put(347,155){\footnotesize .6}
	\put(347,179){\footnotesize .8}\put(349,203){\small 1}
	\put(0,0){\includegraphics[scale=0.28]{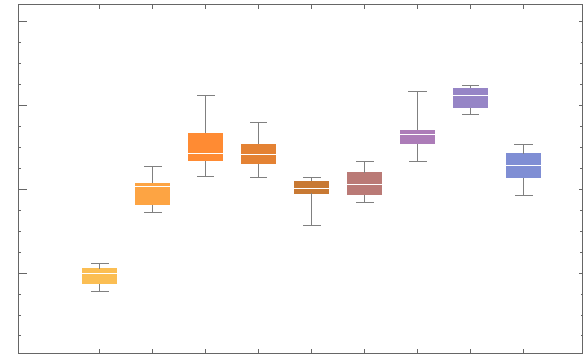}}
	\put(105,10){volunteer 7}
	\put(-3,21){\footnotesize .4}\put(-3,45){\footnotesize .6}
	\put(-3,69){\footnotesize .8}\put(-1,93){\small 1}
	\put(175,0){\includegraphics[scale=0.28]{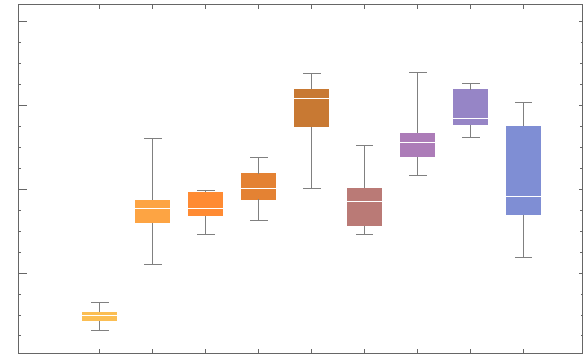}}
	\put(280,10){volunteer 8}
	\put(172,21){\footnotesize .4}\put(172,45){\footnotesize .6}
	\put(172,69){\footnotesize .8}\put(174,93){\small 1}
	\put(350,0){\includegraphics[scale=0.28]{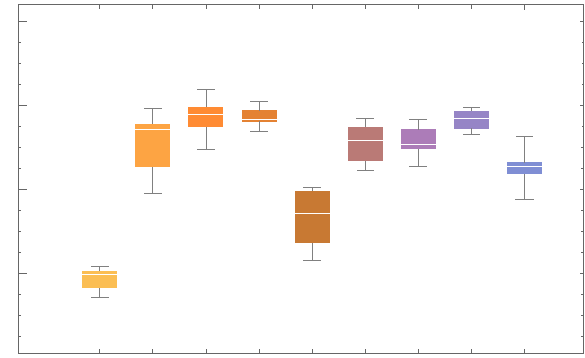}}
	\put(455,10){volunteer 9}
	\put(347,21){\footnotesize .4}\put(347,45){\footnotesize .6}
	\put(347,69){\footnotesize .8}\put(349,93){\small 1}
\end{picture}}
\caption{Statistic of the gait symmetry measurement in our experiments. The horizontal and vertical axes represent respectively gait types and corresponding measurements shown as box and whisker charts. The notation L$|$5cm indicates the simulated gait in which a sole with 5cm of thickness was padded under the left foot, while L$|$4kg means that a 4kg-heavy object was mounted to the left leg, and so on.}
\label{fig:details}
\end{figure*}
The mean values were in the range between 0.30 and 0.44 for normal gaits, and higher measures for the asymmetrical ones. Therefore, considering the returned estimation of an arbitrary gait and that range may allow gait symmetry assessment. However, that range is formed from a set of volunteers, an asymmetrical gait of a subject may thus have an estimation falling inside the normal range of other subjects though this value is still higher than the measure of normal gait with the same subject. This case happened for the R$|$5cm gait of the $4^{th}$ volunteer which was lower than the normal gait of the $6^{th}$ volunteer. Therefore, within-subject analysis should be considered to increase the confidence of the symmetry assessment. Let us see more details of our experimental results in Fig.~\ref{fig:details} instead of only mean values. With most subjects, the measured values tended to decrease when the asymmetry reduces (e.g. L$|$10cm compared with L$|$15cm). This means that our system could be used to assess the recovery of patients after a (knee, hip, etc.) surgery, during a musculoskeletal treatment or after a stroke for instance. In summary, the assessment of gait symmetry can be performed by checking estimated measures with a specific range and confirming the decision based on recent changes of these values (e.g. day by day). Let us notice again that considering only the normal range may not be sufficient since the actual gait symmetry depends on various factors such as health, physical body, and even walking habit. Therefore checking the convergence of symmetry measurements helps us to confirm the normality of patient's gaits.

\subsection{Comparison with other related methods}

\begin{table*}[t]
\centering
\caption{Errors in distinguishing between normal (symmetric) and abnormal (asymmetric) gaits with different approaches}
\label{table:comparison}
\footnotesize
\begin{tabular}{|l|l||l|l|l|l|}
\hline
\textbf{Test subjects} & \textbf{Evaluation} & \textbf{Our method} & \textbf{HMM~\cite{Nguyen2016}} & \textbf{One-class SVM~\cite{Bauckhage2009}} & \textbf{Binary SVM~\cite{Bauckhage2009}} \\ \hline
\multirow{2}{*}{$v_2, v_4, v_7, v_8$} & short-term & 0.042 & 0.335 & 0.227 & 0.157 \\

& full sequence & 0.000 & 0.250 & 0.139 & 0.139 \\ \hline
\multirow{2}{*}{leave-one-out} & short-term & 0.025 ($\pm$ 0.038) & 0.396 ($\pm$ 0.117) & 0.274 ($\pm$ 0.183) & 0.152 ($\pm$ 0.058) \\
& full sequence & 0.000 ($\pm$ 0.000) & 0.198 ($\pm$ 0.250) & 0.136 ($\pm$ 0.070) & 0.111 ($\pm$ 0.000) \\\hline
\multirow{2}{*}{all subjects} & short-term & 0.051 & - & - & - \\
& full sequence & 0.037 & - & - & -\\\hline 
\end{tabular}
\end{table*}

\begin{table*}[!htb]
\centering
\caption{The ability of our method indicated by ROC-based quantities that estimated based on different sizes of cylindrical histogram (evaluated on all subjects)}
\label{table:sizecomparison}
\footnotesize
\begin{tabular}{|c|c||c|c|c|c||c|c|c|c|}
\hline
\multirow{3}{*}{Measure on} & \multirow{3}{*}{Quantity} & \multicolumn{8}{c|}{Histogram size}\\ \cline{3-10} & & \multicolumn{4}{c||}{Increasing of width} & \multicolumn{4}{c|}{Increasing of height} \\ \cline{3-10} 
  & & $16 \times 8$ & $16 \times 16$ & $16 \times 24$ & $16 \times 32$ & $8 \times 16$ & $16 \times 16$ & $24 \times 16$ & $32 \times 16$ \\ \hline\hline
\multirow{2}{*}{Segments}   & AUC & 0.989    & \textbf{0.989} & 0.988 & 0.987 & 0.989 & 0.989 & 0.989 & \textbf{0.989} \\ \cline{2-10} 
 & EER & \textbf{0.043} & 0.050 & 0.044 & 0.044 & \textbf{0.046} & 0.050 &  0.050 & 0.050 \\ \hline
\multirow{2}{*}{Mean} & AUC & \textbf{0.998} & 0.997 & 0.995 & 0.995 & 0.997 & 0.997 & 0.997 & 0.997\\ \cline{2-10} 
 & EER & \textbf{0.014} & 0.028 & 0.028 & 0.028 & \textbf{0.014} & 0.028 & 0.028 & 0.028 \\ \hline
\end{tabular}
\end{table*}
In order to compare the gait-related information gained when exploiting 3D point clouds with other data types, we also performed experiments on the skeletons and silhouettes mentioned in Section~\ref{sec:acquisition}. Method~\cite{Nguyen2016} was employed to deal with the former data type. That study separated an input sequence of skeletons into consecutive gait cycles detected using the distance between two foot joints. A hidden Markov model (HMM) with a specific structure was employed to build a model of normal walking gait cycles as well as to provide a likelihood for each input cycle. The categorization was finally performed by comparing such log-likelihoods with a predefined threshold. For the silhouette input, we used the approach~\cite{Bauckhage2009}, in which the feature extraction was performed on each frame, the temporal context was embedded by vector concatenation, and a support vector machine (SVM) was employed for the task of classification. Both methods aim to classify each input sequence into two categories: normal and abnormal gaits. Their ability was evaluated based on different measures: the Area Under Curve of a Receiver Operating Characteristic (ROC) curve for~\cite{Nguyen2016} and typical classification accuracy for~\cite{Bauckhage2009}. We decided to use the Equal Error Rate (EER) as the measure for comparison because this is estimated according to the ROC curve and its meaning is related to the classification accuracy. Such ROC-based measures have been employed in many problems of binary classification.

The HMM in~\cite{Nguyen2016} was built with only normal gaits. Therefore beside the typical binary SVM, we also modified the model in approach~\cite{Bauckhage2009} to have a one-class SVM. That unsupervised learning is reasonable in practical situations because there are numerous walking gaits that have abnormality, collecting a dataset of such gaits with a high generality is thus difficult. In our experiments, the HMM and one-class SVM were trained with the same dataset consisting of normal gaits of 5 (over 9) subjects ($v_1, v_3, v_5, v_6, v_9$ in Fig.~\ref{fig:overallresult} as suggested in~\cite{NguyenReportDataset}), and the (normal and abnormal) gaits of the remaining subjects were the test set. The binary SVM was also trained on all gaits of those 5 volunteers, and the test set included all gaits of the other 4 volunteers. In order to have a more general evaluation, we also performed the experiments using leave-one-out, i.e. 9-fold cross-validation where each fold contains all 9 gaits of a subject. The assessment was thus represented as mean ($\pm$~std) of the evaluation quantity. The experimental results are presented in Table~\ref{table:comparison}. The notation \textit{short-term} has different meanings: a segment of 120 point clouds in our method, an automatically detected gait cycle in~\cite{Nguyen2016}, and a temporal context of $\Delta = 20$ in~\cite{Bauckhage2009} (i.e. per-frame classification based on vector concatenation of features in 21 recent frames). The notation \textit{full sequence} indicates the classification based on mean values in our work (as shown in Fig.~\ref{fig:overallresult}), lowest averages of log-likelihoods computed on three consecutive cycles in each sequence in~\cite{Nguyen2016}, and alarm triggers on whole input sequences in~\cite{Bauckhage2009}.

According to Table~\ref{table:comparison}, the classification errors resulting from our method are much lower compared with the others. Table~\ref{table:comparison} also shows that in all the 3 methods, the decision provided based on the whole input sequence had a higher confidence compared with short segments. In other words, the mean values in Fig.~\ref{fig:overallresult} were better than individual segment measures in indicating the gait symmetry embedded inside a sequence of point clouds. During our experiments, we observed that the binary SVM~\cite{Bauckhage2009} always classified sequences of normal gaits (according to alarm triggers) into the category of anomaly. This property was clearly showed in the leave-one-out cross validation where the error was 0.111 for all 9 folds. This problem might be due to the large ratio between abnormal and normal gaits (8:1), and a binary (i.e. supervised) SVM was thus not really appropriate for the task of detecting abnormal gaits where there are numerous types of abnormal walking. It was also noticeable that the approach~\cite{Nguyen2016} could be improved to get better results by modifying the width of sliding window since the frame rate of our data acquisition was lower than the system in~\cite{Nguyen2016}.

\subsection{Sensitivity to size of cylindrical histogram}\label{sec:sizeaffection}
The cylindrical histogram plays the main role in our approach and also affects the gait symmetry assessment. By changing the histogram's size, i.e. number of sectors, the range of mean values in Section~\ref{sec:ourresult} would be different. The ability of distinguishing two gait types would also change. We can guess that a histogram with small resolution can reduce the computational cost of the entire system but may not have enough details for describing body postures. On the contrary, using a histogram formed from a large number of cylindrical sectors may also reduce the system's efficiency. In that case, each sector covers a small volume with low numbers of 3D points, the result of eq.~(\ref{eq:crosscorrelation}) is thus sensitive to noise in the input 3D point clouds. In summary, the system accuracy can be improved by a careful selection of histogram size. Table~\ref{table:sizecomparison} shows the abilities (according to AUCs and EERs of ROC curves) of our system for various histogram resolutions in distinguishing symmetrical and asymmetrical walking gaits. In this table, we focus on the mean-based measurement because it describes the gait symmetry better than segments (according to Table~\ref{table:comparison}). The ability of our method tended to reduce, i.e. increasing of EER and decreasing of AUC, when we set a high value for the histogram width. The height of cylindrical histograms had a lower effect since the AUC and EER (for both segments and means) were almost unchanged when the height exceeded a particular threshold.

\section{Conclusion}\label{sec:conclusion}
In this paper, we have presented an original and efficient low-cost system for assessing gait symmetry using a ToF depth camera together with two mirrors. The input of the proposed method is a sequence of 3D point clouds representing the subject's postures when walking on a treadmill. By fitting a cylinder on each point cloud, a cylindrical histogram is formed to describe the corresponding gait in the manner of self-symmetry. Cross-correlation is then applied on each pair of sequences of half-body sub-histograms to measure the gait symmetry along the movement. The ability of our method has been demonstrated via a dataset of 9 subjects and 9 gait types. Our approach also outperforms some related works, that employ skeletons and frontal view silhouettes as the input, in the task of distinguishing normal (symmetric) and abnormal (asymmetric) walking gaits. The resulting system is thus a promising tool for a wide range of clinical applications by providing relevant gait symmetry information. Patient screening, follow-up after surgery, treatment or assessing recovery after a stroke are obvious applications that come to mind. As future work, the proposed method will be modified focusing on particular pathological gaits such as diplegic, hemiplegic, choreiform, and Parkinsonian~\cite{Neurologic} in order to support the gait diagnosis on patients.

\subsubsection*{Acknowledgment}
The authors would like to thank the NSERC (Natural Sciences and Engineering Research Council of Canada) for having supported this work (Discovery Grant RGPIN-2015-05671).

\bibliography{references}
\bibliographystyle{iclr2019_conference}

\end{document}